\documentclass[10pt,twocolumn,letterpaper]{article}

\PassOptionsToPackage{table}{xcolor}
\usepackage[pagenumbers]{cvpr} 

\usepackage{multirow}
\usepackage{makecell}
\usepackage{bm}
\usepackage{xspace}
\usepackage{pifont}  

\newcommand{\TODO}[1]{\textbf{\color{red}[TODO: #1]}}
\renewcommand{\TODO}[1]{}

\newcommand{\paper}{PanoEvo\xspace}
\newcommand{\data}{PanoEvo-Data\xspace}
\newcommand{\evolver}{Panoramic Evolver\xspace}
\newcommand{\renderer}{Perspective Renderer\xspace}
\newcommand{\E}{\mathcal{E}}
\newcommand{\R}{\mathcal{R}}

\definecolor{firstbg}{RGB}{254,234,234}
\definecolor{secondbg}{RGB}{231,240,253}
\definecolor{oursbg}{RGB}{246,246,246}
\newcommand{\first}[1]{\cellcolor{firstbg}\textbf{#1}}
\newcommand{\second}[1]{\cellcolor{secondbg}\underline{#1}}

\definecolor{cvprblue}{rgb}{0.21,0.49,0.74}
\usepackage[pagebackref,breaklinks,colorlinks,allcolors=cvprblue]{hyperref}
\ifdefined\pdfsuppressptexinfo\pdfsuppressptexinfo=-1\fi

\def\confName{CVPR}
\def\confYear{2027}

\title{Evolve in Panorama, Render in View:\\Panoramic World State Evolution for Video World Models}

\author{
Longtao Jiang$^{1}$\quad Jianxiang He$^{1}$\quad Peerawat Chomphooyod$^{1}$\quad Mingfei Han$^{1}$\quad Jianmin Bao$^{2}$\quad Tingjun Dai$^{1}$\quad \\ Jinjie Wei$^{1}$\quad
Zhiran Zhang$^{1}$\quad Xin Tao$^{2}$\quad Zhihui Li$^{1}$\quad Salman Khan$^{1}$\quad Xiaojun Chang$^{1}$\\[0.4em]
{\normalsize $^{1}$Mohamed bin Zayed University of Artificial Intelligence\qquad $^{2}$Kuaishou Technology}
}

\begin{document}

\twocolumn[{%
\renewcommand\twocolumn[1][]{#1}%
\maketitle
\vspace{-24pt}
\begin{center}
  \centering
  \includegraphics[width=\textwidth]{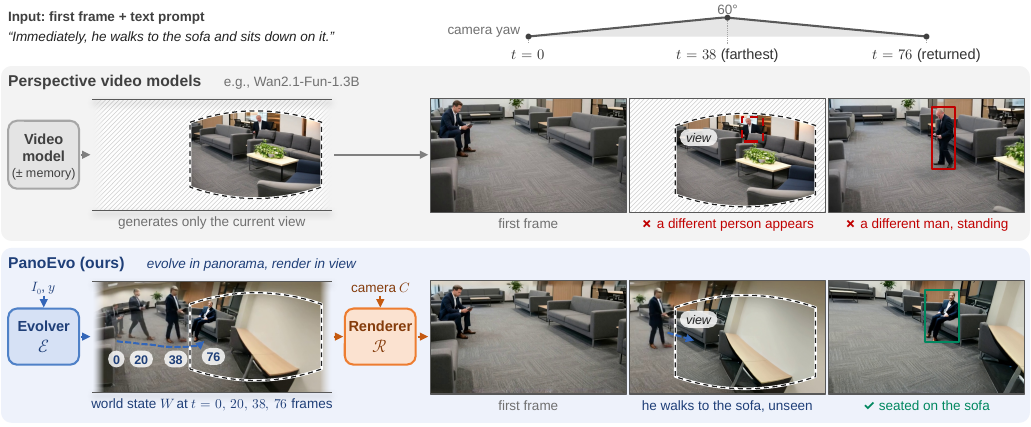}
  \vspace{-1.5em}
  \captionof{figure}{\textbf{Evolve in panorama, render in view.}
  The camera turns $60^\circ$ away from a man asked to walk to the sofa and sit down, and returns.
  Top: perspective video models such as Wan2.1-Fun-1.3B~\cite{videoxfun} generate only the current view and show a different man on return.
  Bottom: \paper evolves the world state $W$ in a panoramic latent, where the man reaches the sofa unseen, and renders views from $W$.}
  \label{fig:teaser}
\end{center}
}]

\begin{abstract}
Video world models should maintain a world that keeps evolving as the viewpoint changes.
Yet when the camera turns away and returns, current models often freeze the unobserved scene, lose objects, or change their appearance, because content outside the view has no representation that keeps changing.
We present \paper, which uses a \evolver to evolve the world state in the latent space of a panoramic video diffusion model, where every viewing direction is present at every time step, so the scene evolves whether or not it is observed.
Existing panoramic video models are not trained to evolve a scene from a single image or the objects within it, so we build \data, an event-centric dataset from real 360$^\circ$ videos, and train the \evolver coarse-to-fine as an image-to-panoramic-video model.
Since each view covers only a small part of the low-resolution panorama, directly projected views are blurry and often show low-quality details or structural errors.
A \renderer, built on a perspective video diffusion model, reads this state via latent-space projection and direction-restricted cross-attention to render sharp, high-quality views.
\paper achieves the best overall score among 23 publicly evaluated models on WRBench and the highest total score on STEVO-Bench.
Code, models, and data will be released.
\end{abstract}

\section{Introduction}
\label{sec:intro}

Video world models generate future observations of a scene from an initial image and a sequence of viewpoints or actions~\cite{genie3,worldplay,lingbotworld,cosmos}.
As they move toward interactive exploration, they are expected to maintain a \emph{persistent} world.
Persistence, however, is more than geometric consistency.
The world keeps evolving: a person keeps walking, a dog keeps playing, and an action that has started keeps unfolding, whether or not the camera looks at it.
Recent benchmarks test this by turning the camera away and back~\cite{wrbench,stevo,memobench}, and show that current models fall short.
On return, the unobserved scene is frozen in its last observed state, objects vanish, or their appearance changes.

The root cause lies in what a perspective video model represents.
It models the content inside the current view, and once content leaves it, nothing in the model represents it, let alone advances it.
Existing remedies mainly retain what has been seen.
Geometry caches and 3D memories~\cite{gen3c,versecrafter,spatia,evoworldpano} keep the scene geometry, source-video re-rendering~\cite{recammaster,inspatio} reuses observed frames, and memory or state mechanisms~\cite{hydra,liveworld,remind,evoworldcausal} retrieve past observations or reason about entities explicitly.
These approaches preserve how the world looked when observed, but few densely represent unobserved content that keeps changing.

We observe that, for a rotating camera, a panoramic video latent provides exactly such a representation.
Every viewing direction is present at every time step, so a panoramic video diffusion model evolves the whole scene jointly in its latent space, regardless of where the camera points, and any requested view is a projection of this state.
Directly projected views, however, are poor.
At a typical $512{\times}1024$ resolution, a $90^\circ$ view covers only about 256 pixels horizontally and must be upsampled over three times to $480{\times}832$, blurring the view and degrading object structures.
We therefore assign the two roles to two models (\cref{fig:teaser}).
A panoramic video diffusion model maintains the evolving world state, and a perspective video diffusion model reads it to render high-quality views.
We call this framework \paper: \emph{evolve in panorama, render in view}.

Realizing this idea requires two components.
First, the panoramic model has to work like a conventional image-to-video model, the common interface of video world models~\cite{genie3,worldplay,inspatio}: given a single image, it should evolve the whole scene.
Existing panoramic video models are text-driven~\cite{dvd360,panowan,viewpoint} or complete panoramas from video observations~\cite{panoworld,imagine360}, panoramic world models start from a full panoramic image~\cite{evoworldpano}, and how individual objects evolve remains underexplored.
Zero-shot, a strong panoramic model completes the scene but evolves it poorly (\cref{sec:exp_ablation}).
We build \data from real 360$^\circ$ videos and train the \evolver coarse-to-fine.
A \emph{Completion stage} on about 20K curated clips teaches image-conditioned panoramic completion, and an \emph{Evolution stage} on 1.4K verified event clips teaches object-level state evolution.
Second, the \renderer has to read the state faithfully and improve visual quality.
Built on a perspective video model, it projects the panoramic latent into each view in latent space, reads further context through frustum cross-attention, and is guided by an auxiliary read-out loss.

\paper achieves the best overall score on WRBench among 23 publicly evaluated models (.753 vs.\ .741 for the strongest baseline), together with the best scores on visible state and both re-observation metrics.
On STEVO-Bench, it attains the highest total score (364.7 vs.\ 300.9).
Compared with direct projection of the panorama, the \renderer raises VBench imaging quality from .538 to .729 while largely preserving the evolved states.

Our contributions are threefold:
\begin{itemize}[leftmargin=*,itemsep=1pt,topsep=2pt]
  \item We propose \paper, which evolves the world state in the latent space of a panoramic video diffusion model and renders views from it with a perspective one.
  \item We build \data from real 360$^\circ$ videos and train the \evolver coarse-to-fine to evolve object-level states from a single image.
  \item We design a \renderer that reads the state through latent-space projection and frustum cross-attention. \paper is state of the art on both benchmarks.
\end{itemize}

\section{Related Work}
\label{sec:related}

\begin{figure*}[t]
  \centering
  \includegraphics[width=\textwidth]{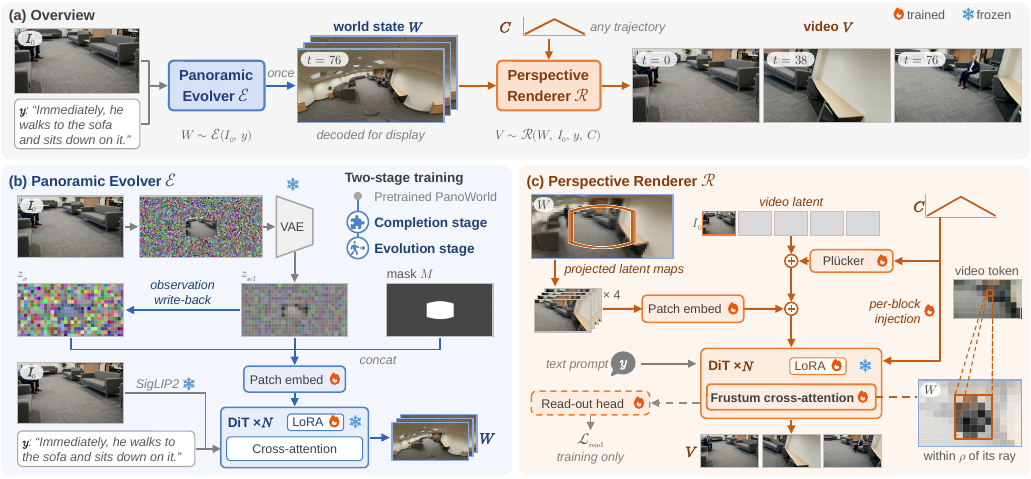}
  \vspace{-1.5em}
  \caption{\textbf{Overview of \paper.}
  (a) $\E$ turns an image $I_0$ and optional text $y$ into an evolving panoramic latent $W$, the world state, from which $\R$ renders a video $V$ along any trajectory $C$.
  (b) $\E$ is fine-tuned in two stages and writes observed cells back from $z_{\text{ref}}$.
  (c) $\R$ reads $W$ via projected latent maps and frustum cross-attention, where each video token attends to panoramic tokens within $\rho$ of its ray.}
  \vspace{-1em}
  \label{fig:pipeline}
\end{figure*}

\paragraph{Video world models with memory and state.}
Interactive and camera-controlled video world models~\cite{genie3,worldplay,gamecraft,lingbotworld,magicworld,cosmos} generate long explorations from an initial image.
Camera control is usually injected into video diffusion models~\cite{wan,hunyuanvideo,cogvideox} through Pl\"ucker embeddings~\cite{plucker,cameractrl,motionctrl,ac3d}, with open-source variants available~\cite{videoxfun}.
To add memory, geometry-based methods maintain point clouds or 3D caches and condition generation on their renderings~\cite{gen3c,versecrafter,spatia,evoworldpano}.
Source-video methods re-render a given video along a new trajectory~\cite{recammaster,inspatio}.
These methods keep revisited regions geometrically consistent but treat dynamic content as static or remove it.
Closer to our goal, HyDRA~\cite{hydra} tracks dynamic subjects over a static memory, LiveWorld~\cite{liveworld} advances dynamic entities with an evolution engine, and ReMind~\cite{remind} recovers evolving states across interruptions.
Other works reason about out-of-view events with a vision-language model~\cite{evoworldcausal} or control objects along trajectories planned by a language model~\cite{worlddirector}.
Recent evaluations expose the remaining gap through camera look-away, occlusion, and illumination changes~\cite{wrbench,stevo,memobench}, where unobserved content is often frozen, lost, or altered.
Unlike these methods, \paper keeps a dense panoramic latent of the whole scene that evolves at every time step, so unobserved content advances without explicit entity modeling, memory retrieval, or event reasoning.

\paragraph{Panoramic video generation.}
Panoramic video models generate 360$^\circ$ videos from text~\cite{dvd360,panowan,viewpoint}, outpaint perspective videos to panoramas~\cite{imagine360}, use cube-map representations for high resolution~\cite{cubecomposer}, or support roaming and world modeling~\cite{omniroam,panoworld,moverse,evoworldpano}, and panoramic image models~\cite{dit360} provide static first frames.
PanoWorld~\cite{panoworld} treats the panoramic latent as a geometrically consistent world representation, and we build the \evolver on its released weights.
These works focus on the panorama itself and its spatial consistency.
We instead use a panoramic model as the state of a perspective world model, start from a single image, and target object-level evolution.

\section{Method}
\label{sec:method}

\subsection{Problem Setting and Overview}
\label{sec:overview}

Given a perspective frame $I_0$, an optional text prompt $y$, and a camera trajectory $C=\{R_t\}_{t=0}^{T-1}$, we generate a perspective video $V=\{I_t\}_{t=0}^{T-1}$ that starts at $I_0$ and follows $C$.

\paragraph{Why a panoramic state.}
For a rotating camera, every perspective frame is a deterministic function of the panoramic frame at the same time,
\begin{equation}
  I_t = \Pi(P_t, R_t),
  \label{eq:projection}
\end{equation}
where $P_t$ is the $360^\circ$ frame at time $t$ and $\Pi$ is the perspective projection.
Content outside the current view is not missing, only not projected.
Generating $P_{0:T-1}$ from $I_0$ amounts to evolving the whole world, and the trajectory only selects which part is shown.
A perspective model, in contrast, has to decide what an unobserved region looks like when it re-enters the view, without having represented it in between, which is where frozen scenes and vanishing objects arise.

\paragraph{Factorization.}
\paper factorizes generation through a panoramic world state $W$, generated by the \evolver $\E$ and rendered by the \renderer $\R$ (\cref{fig:pipeline}):
\begin{equation}
  W \sim \E(I_0, y), \qquad V \sim \R(W, I_0, y, C),
  \label{eq:factor}
\end{equation}
where $W$ is the latent of an equirectangular (ERP) panoramic video under the Wan2.1 VAE~\cite{wan}, the latent space that $\R$ reads.
$\E$ determines \emph{what happens} in the world, including outside the view, and $\R$ reads it faithfully along $C$ and renders coherent, high-quality views.
$\E$ and $\R$ are trained separately, and at inference $\E$ generates $W$ once and $\R$ renders any number of views from it.

\subsection{Panoramic Evolver}
\label{sec:evolver}

\paragraph{Conditioning and observation write-back.}
We build $\E$ on PanoWorld~\cite{panoworld}, a panoramic video diffusion transformer based on Cosmos~\cite{cosmos}, and reuse its native completion pathway without adding input channels.
We place $I_0$ at the front of the ERP canvas (yaw $0^\circ$, pitch $0^\circ$) to obtain a sparse observation $X_{\text{obs}}$ and its visibility mask $M$.
Unobserved pixels are filled with Gaussian noise, $X_{\text{cond}} = M\odot X_{\text{obs}} + (1-M)\odot \epsilon_{\text{fill}}$ with $\epsilon_{\text{fill}}\sim\mathcal{N}(0,0.5^2 I)$, and $X_{\text{cond}}$ is encoded into a reference latent $z_{\text{ref}}$, which is concatenated with the latent-resolution mask along the channel dimension.
SigLIP2~\cite{siglip2} features of $I_0$ enter through cross-attention, and the text $y$ is encoded by Cosmos-Reason1~\cite{cosmosreason}.
The condition is encoded independently of the target panorama, so it never contains unobserved content.
Before every network evaluation, the fully observed latent cells $\bar M$ (coverage above 0.99) are replaced by the clean $z_{\text{ref}}$ and excluded from the loss, and the replacement is applied once more after sampling.
Partially covered cells are generated and supervised.
This native write-back is essential for single-image inputs, since without it the fine-tuned model learns to ignore the observed image.

\paragraph{Objective.}
Let $z_0$ be the latent of the target panoramic video.
Following rectified flow~\cite{rectifiedflow,flowmatching}, we sample a noise level $\sigma$ with logit-normal sampling and a shift of 5~\cite{sd3}, form $z_\sigma = \sigma\epsilon + (1-\sigma)z_0$ with $\epsilon\sim\mathcal{N}(0,I)$, and minimize
\begin{equation}
  \begin{split}
  r_\theta &= v_\theta(z_\sigma,\sigma,c) - (\epsilon - z_0),\\
  \mathcal{L}_{\E} &= \mathbb{E}_{\sigma,\epsilon}\big[\, \|(1-\bar M)\odot r_\theta\|^2 \big],
  \end{split}
  \label{eq:evolver_loss}
\end{equation}
where $v_\theta$ is the velocity predicted by $\E$, $r_\theta$ its residual, and $c$ the conditions ($z_{\text{ref}}$ with its mask, $I_0$, and $y$).

\paragraph{Coarse-to-fine training.}
We train $\E$ in two stages on the subsets of \data (\cref{sec:data}).
The \emph{Completion stage} adapts the model to image conditions on the Coarse subset.
Besides single-image conditions, it uses trajectory-video conditions as an auxiliary task, in which observations along a random yaw trajectory are back-projected to the canvas, so that the model sees content entering and leaving the view.
The \emph{Evolution stage} starts from the Completion-stage weights and trains on the Event subset with image conditions only.
Its prompts are detailed, brief, or empty, so that the model learns to evolve the scene under different amounts of textual guidance.
Completion needs scale but tolerates loosely matched text, whereas object-level evolution needs verified events and matching prompts, which limits the amount of data.
Starting from the Completion stage lets the Evolution stage focus on how objects change rather than on how to complete a panorama (\cref{tab:stages}).

\subsection{Perspective Renderer}
\label{sec:renderer}

\begin{figure*}[t]
  \centering
  \includegraphics[width=\textwidth]{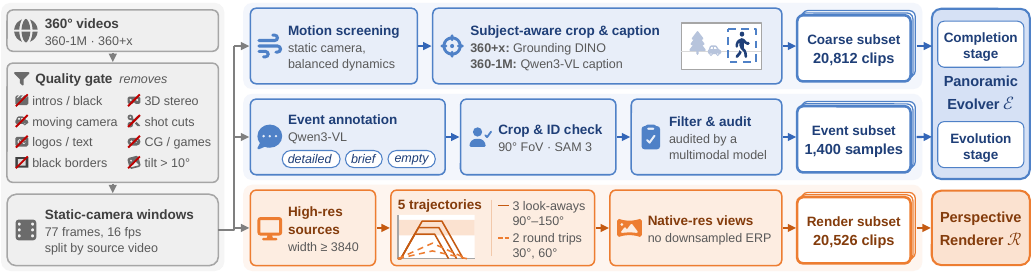}
  \vspace{-1.5em}
  \caption{\textbf{Construction of \data.} A shared quality gate feeds three branches, two for the stages of $\E$ and one for $\R$.}
  \label{fig:data}
  \vspace{-1em}
\end{figure*}

$\R$ is built on Wan2.1-T2V-1.3B~\cite{wan} and renders the view from $W$, $I_0$, $y$, and $C$.
The trajectory enters through Pl\"ucker embeddings~\cite{plucker,cameractrl} at the input and a zero-initialized per-block camera injection, and $I_0$ replaces the first latent frame.
$\R$ reads $W$ through two complementary pathways.

\paragraph{Projected latent maps.}
Each latent frame $k$ of the video corresponds to a group $G_k$ of four RGB frames (one for $k{=}0$).
For each frame $j\in G_k$, we cast the rays of the perspective latent grid, rotate them by $R_j$, and bilinearly sample the panoramic latent frame $W_k$:
\begin{equation}
  \begin{split}
  \tilde W_k^{(j)}(u) &= W_k\big(\pi_{\text{ERP}}(R_j\, \hat d(u))\big),\\
  \hat d(u) &= K^{-1}\tilde u \,/\, \|K^{-1}\tilde u\|,
  \end{split}
  \label{eq:proj}
\end{equation}
where $u$ is a perspective latent grid location, $\tilde u$ the homogeneous coordinate of its center, $\hat d(u)$ its unit ray in camera coordinates, $K$ the intrinsics at latent resolution, and $\pi_{\text{ERP}}$ maps a unit direction to panoramic latent coordinates.
The four maps are stacked along channels and added to the video tokens by a zero-initialized patch embedding.
The projection needs no decoding, and per-frame rotation aligns it with every RGB frame.
It gives every output token a pixel-aligned estimate of the panorama in its direction, which anchors the view geometry under large rotations.

\paragraph{Frustum cross-attention.}
To aggregate panoramic context beyond this pixel-aligned estimate, every block contains a cross-attention from video tokens to panoramic tokens $p$, obtained by patchifying $W$ and adding a Fourier encoding of their center directions $d_p$.
A query token $u$ of latent frame $k$ attends only to the set $\mathcal{N}_k(u)$ of panoramic tokens of the same latent frame that lie within an angular radius $\rho{=}15^\circ$ of its viewing direction in any frame of $G_k$:
\begin{equation}
  \mathcal{N}_k(u) = \Big\{\, p : \min_{j\in G_k} \angle\big(d_p,\, R_j\, \hat d(u)\big) \le \rho \,\Big\}.
  \label{eq:frustum}
\end{equation}
where $\angle$ is the angle between unit vectors.
Defined on the sphere, the window covers a consistent solid angle at all latitudes despite the ERP distortion.
The attention output passes through a zero-initialized channel-wise gate, so $\R$ starts as the original model.
A linear read-out head $h_b$ on each frustum block $b\in\mathcal{B}$ maps the attended features $a_b$ to latent channels,
\begin{equation}
  \mathcal{L}_{\text{read}} = \frac{1}{|\mathcal{B}|}\sum_{b\in\mathcal{B}} \big\| h_b(a_b) - \mathrm{D}(z_0) \big\|^2,
  \label{eq:read}
\end{equation}
where $z_0$ is the target view latent and $\mathrm{D}$ is $2{\times}2$ average pooling to tokens.
It encourages the attention to retrieve panoramic content rather than stay inactive behind its gate.

\paragraph{Training conditions.}
Following cascaded generation practice~\cite{cdm,imagen}, $\R$ is trained on ground-truth panoramic latents and views projected from the same high-resolution panorama, and reads the generated panorama at inference.
During training, the condition is perturbed as $W\leftarrow(1-s)W+s\,\epsilon$ with a random level $s$, which is embedded and added to the timestep embedding, and $s{=}0$ at inference.
The perturbation teaches $\R$ to treat the panorama as an estimate to be refined rather than a signal to be copied, since generated panoramas are softer than ground truth.
The panorama and the text are also dropped at random.

\paragraph{Objective.}
Following the flow-matching parameterization of the base model~\cite{wan,flowmatching}, we form $z_\sigma = \sigma\epsilon + (1-\sigma)z_0$ and minimize
\begin{equation}
  \begin{split}
  r_\phi &= v_\phi(z_\sigma,\sigma,c') - (\epsilon - z_0),\\
  \mathcal{L}_{\R} &= \mathbb{E}_{\sigma,\epsilon}\big[\, \omega(\sigma)\, \|(1-m_0)\odot r_\phi\|^2 \big] + \lambda\,\mathcal{L}_{\text{read}},
  \end{split}
  \label{eq:renderer_loss}
\end{equation}
where $v_\phi$ is the velocity predicted by $\R$, $r_\phi$ its residual, $c'=(W,I_0,y,C)$, $m_0$ a mask that excludes the first latent frame, which holds $I_0$, from the loss, $\omega(\sigma)$ a Gaussian noise-level weight centered at $\sigma{=}0.5$, and $\lambda{=}0.1$.

\section{\data}
\label{sec:data}

Both models are trained on real panoramic videos with a fixed camera center, which we build from 360-1M~\cite{m360} and 360+x~\cite{x360} (\cref{fig:data,tab:data}).
All clips are 77-frame, 16\,fps windows of continuous static-camera footage that pass a shared quality gate.
Three principles guide the construction.
(i) \emph{Real footage only}, avoiding the domain gap of simulated scenes.
(ii) \emph{Scale where it suffices, verification where it matters.} The Coarse and Render subsets favor scale with inexpensive rules, while the Event subset admits only samples with verified events.
(iii) \emph{Supervision matched to each model.} The \evolver is supervised by complete panoramas at its world-state resolution, and the \renderer by views projected from native-resolution sources.
Details are in the supplementary material.

\paragraph{Coarse subset.}
The Coarse subset teaches image-conditioned completion at scale, with inexpensive rules against three failure cases.
\emph{Camera motion} is measured from the global panoramic flow.
\emph{Freezing} is limited by capping static and low-motion windows by their dynamic area.
\emph{Duplicated subjects} arise when the text mentions a subject outside the cropped first frame, which teaches the model to generate it elsewhere.
For 360+x, whose text is a scene and action label, the crop is oriented toward the subject detected by Grounding DINO~\cite{groundingdino}.
360-1M has no text, so its crop is oriented by motion and Qwen3-VL~\cite{qwen3vl} captions the crop, naming only subjects that Grounding DINO detects in it.

\paragraph{Event subset.}
The Event subset teaches object-level evolution with verified events.
Qwen3-VL annotates each candidate on multi-directional previews, using the video as evidence and the original caption only as a hint, and records the event, the entities to check and their expected changes, those required in the first frame, and the timing.
Each verified event yields up to three prompts of decreasing detail.
For a person who walks to a bench and sits down, the detailed prompt describes the walk and the sitting, the brief prompt only says that the person heads toward the bench, and the sample without a prompt starts once the walk is visible in the first frame.
The first frame shows the required entities within a $90^\circ$ field of view, and the target is verified against SAM\,3~\cite{sam3} tracks.
We further reject close-ups, moving cameras, and prompt--video mismatches, and audit every remaining sample with Claude Opus 5.5.
The final 1,400 samples from 846 source videos cover in-place actions (42\%), walking toward a target (26\%), holding a state (25\%), and walking to then interacting with an object (6\%).

\begin{table}[t]
  \centering
  \caption{\textbf{\data.} Three subsets from real 360$^\circ$ videos, split by source video and sharing no time windows.}
  \vspace{-0.5em}
  \label{tab:data}
  \setlength{\tabcolsep}{3pt}
  \resizebox{\linewidth}{!}{%
  \begin{tabular}{@{}lccc@{}}
    \toprule
     & Coarse & Event & Render \\
    \midrule
    Clips & 20,812 & 1,400 & 20,526 ($\times$5 traj.) \\
    Source videos & -- & 846 & -- \\
    Resolution & ERP $512{\times}1024$ & ERP $512{\times}1024$ & native ($\ge$3840 wide) \\
    Text & label / VLM caption & detailed / brief / empty & label / empty \\
    Annotation & rule-based & events, verified & trajectories \\
    Used for & $\E$, Completion & $\E$, Evolution & $\R$ \\
    \bottomrule
  \end{tabular}}
  \vspace{-1.5em}
\end{table}

\paragraph{Render subset.}
The Render subset supervises $\R$ at high resolution.
Since a $90^\circ$ view at $832$ pixels needs a 3,300-pixel-wide ERP, we keep only sources at least 3,840 pixels wide.
Each of the 20,526 clips has five trajectories, three look-aways ($90^\circ$--$150^\circ$) and $30^\circ$ and $60^\circ$ round trips.
Target views are projected from the native-resolution source to $480{\times}832$, while conditions are $512{\times}1024$ ERP latents.
Consecutive samples often share a clip but differ in trajectory, so only the camera tells $\R$ which part to render.

\begin{table*}[t]
  \centering
  \caption{\textbf{Results on WRBench.}
  Baselines are grouped by viewpoint condition as in~\cite{wrbench}.
  Params: backbone size (total for MoE); for \paper, both DiTs with added modules.
  Content Avg.\ averages Integ.\ and the four spatial/state metrics, and Spatial--State Avg.\ only the latter.
  The other nine officially evaluated models are low-sample and listed in the supplement.
  Best and second best: \colorbox{firstbg}{\textbf{red}} and \colorbox{secondbg}{\underline{blue}}.}
  \vspace{-0.5em}
  \label{tab:wrbench}
  \setlength{\tabcolsep}{4pt}
  \resizebox{0.98\textwidth}{!}{%
  \begin{tabular}{lcccccccccc}
    \toprule
    & & & & \multicolumn{2}{c}{Visible} & \multicolumn{2}{c}{Re-observed} & & Content & Spatial-- \\
    \cmidrule(lr){5-6}\cmidrule(lr){7-8}
    Model & Params & Cam Align. & Integ. & Spatial & State & Spatial & State & Avg. & Avg. & State Avg. \\
\midrule
\multicolumn{11}{l}{\textit{Source-video condition}}\\
InSpatio-World 14B~\cite{inspatio} & 14B & \second{.980} & .700 & .769 & .624 & \second{.729} & \second{.643} & \second{.741} & \second{.693} & \second{.691} \\
ReCamMaster~\cite{recammaster} & 1.3B & .937 & .623 & .653 & .493 & .662 & .609 & .663 & .608 & .604 \\
HyDRA~\cite{hydra} & 1.3B & .855 & .611 & .587 & .460 & .468 & .410 & .565 & .507 & .481 \\
\midrule
\multicolumn{11}{l}{\textit{Geometry-cache condition}}\\
Gen3C~\cite{gen3c} & 7B & \first{.993} & .578 & .677 & .535 & .681 & .604 & .678 & .615 & .624 \\
VerseCrafter~\cite{versecrafter} & 14B & .901 & .743 & .669 & .483 & .610 & .551 & .660 & .611 & .578 \\
Spatia~\cite{spatia} & 5B & .682 & .669 & .688 & .515 & .608 & .557 & .620 & .607 & .592 \\
\midrule
\multicolumn{11}{l}{\textit{Prompt-only condition}}\\
Wan2.6 I2V~\cite{wan} & API & .121 & \first{.795} & \first{.810} & \second{.645} & .653 & .526 & .592 & .686 & .658 \\
\midrule
\multicolumn{11}{l}{\textit{Model-inferred condition}}\\
Wan2.2-Fun-A14B~\cite{videoxfun} & 27B & .762 & .737 & .752 & .576 & .697 & .628 & .692 & .678 & .663 \\
Wan2.1-Fun-1.3B~\cite{videoxfun} & 1.3B & .877 & .772 & .665 & .474 & .699 & .628 & .686 & .648 & .616 \\
LiveWorld~\cite{liveworld} & 14B & .977 & .611 & .652 & .501 & .674 & .610 & .671 & .610 & .609 \\
Wan2.2-Fun-5B~\cite{videoxfun} & 5B & .612 & .695 & .762 & .576 & .719 & .633 & .666 & .677 & .672 \\
Wan2.1-Fun-14B~\cite{videoxfun} & 14B & .740 & .758 & .675 & .487 & .657 & .601 & .653 & .636 & .605 \\
Hunyuan WorldPlay~\cite{worldplay} & 8B & .534 & \second{.793} & .679 & .486 & .651 & .588 & .622 & .640 & .601 \\
MagicWorld~\cite{magicworld} & 1.3B & .849 & .496 & .573 & .432 & .588 & .571 & .585 & .532 & .541 \\
\rowcolor{oursbg}\textbf{\paper (ours)} & 2.4B+1.8B & .950 & .720 & \second{.799} & \first{.651} & \first{.754} & \first{.646} & \first{.753} & \first{.714} & \first{.712} \\
    \bottomrule
  \end{tabular}}
  \vspace{-1em}
\end{table*}

\section{Experiments}
\label{sec:exp}

\subsection{Experimental Setup}
\label{sec:exp_setup}

\paragraph{Benchmarks and metrics.}
\emph{WRBench}~\cite{wrbench} tests whether the spatial layout and object states stay consistent when hidden by camera rotation and re-observed.
It covers static views and yaw rotations of two magnitudes in both directions, and reports camera alignment (Cam Align.), integrity (Integ.), visible and re-observed spatial and state consistency, and their average (Avg.).
We re-score all baselines (\cref{tab:wrbench}) with the official code on their released videos, since it differs from that behind the published table.
\emph{STEVO-Bench}~\cite{stevo} tests whether an ongoing process continues unobserved to a later state.
We report its five metrics and their sum (Total) as in ReMind~\cite{remind}.
\emph{VBench}~\cite{vbench,vbenchpp} measures visual quality on WRBench videos with eight dimensions of its image-to-video mode (excluding camera motion), plus CLIP-IQA+~\cite{clipiqa} and Laplacian sharpness.

\paragraph{Implementation details.}
We generate 77 frames at 16\,fps and $480{\times}832$, placing $I_0$ with a nominal $90^\circ$ horizontal field of view, and $W\in\mathbb{R}^{16\times 20\times 64\times 128}$ is the latent of a $512{\times}1024$ ERP video.
$\E$ is fine-tuned from PanoWorld with rank-64 LoRA~\cite{lora}, for 3 epochs in the Completion stage (batch 32, constant learning rate $10^{-4}$, single-image and trajectory-video conditions at 0.8:0.2) and 11 epochs in the Evolution stage (batch 8, cosine schedule).
$\R$ is fine-tuned from Wan2.1-T2V-1.3B with rank-64 LoRA and fully trained new modules for 5 epochs (batch 8).
The condition is perturbed with probability 0.8 and $s\sim\mathcal{U}(0,0.5)$, and the panorama and the text are each dropped with probability 0.1.
$\E$ samples with 35 steps and CFG 7.0, and $\R$ with 50 steps and text CFG 5.0.
For each test case, $\E$ generates one panorama from $I_0$ and the benchmark text without prompt rewriting, and $\R$ renders every camera setting from it.
Full settings are in the supplementary material.

\begin{figure*}[t]
  \centering
  \includegraphics[width=\textwidth]{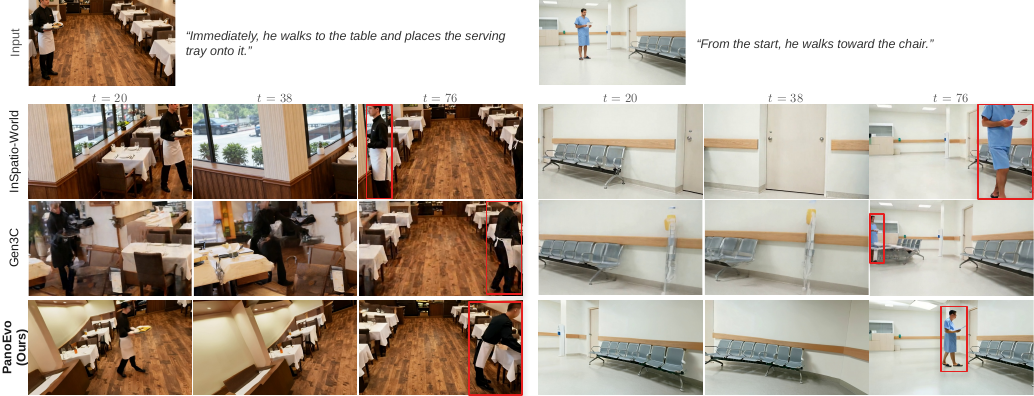}
  \vspace{-1.5em}
  \caption{\textbf{Qualitative comparison.} Views at $t{=}20$ (turning away), 38 (farthest), and 76 (returned), with the subject boxed.}
  \label{fig:qualitative}
  \vspace{-1em}
\end{figure*}

\subsection{Main Results}
\label{sec:exp_main}

\paragraph{WRBench.}
As shown in \cref{tab:wrbench}, \paper attains the best official average (.753, vs.\ .741 for InSpatio-World 14B) and the best visible-state, re-observed spatial, and re-observed state scores, as well as both content averages.
Its camera alignment (.950) is close to that of the strongest geometry-cache and source-video models, and its two diffusion transformers (2.4B and 1.8B, including our added modules) are smaller than the backbones of most competitive baselines.
The prompt-only Wan2.6 I2V obtains high integrity and visible-spatial scores, but its camera alignment of .121 shows that the camera barely turns, so little content leaves the view and these scores are inflated.
Among models conditioned only on the input frame and the trajectory, \paper improves the official average by .061 over the strongest one.

\begin{table}[t]
  \centering
  \caption{\textbf{Results on STEVO-Bench} (\%). Baseline scores are the official leaderboard values reported in ReMind~\cite{remind}.}
  \vspace{-0.5em}
  \label{tab:stevo}
  \setlength{\tabcolsep}{2.6pt}
  \resizebox{0.98\linewidth}{!}{%
  \begin{tabular}{lcccccc}
    \toprule
    Model & \makecell{State\\Progress} & \makecell{Physical\\Plaus.} & Coherence & \makecell{Obs.\\Control} & \makecell{Action\\Control} & Total \\
\midrule
\multicolumn{7}{l}{\textit{Video models}}\\
HunyuanVideo 1.5~\cite{hunyuanvideo15} & 4.1 & 42.1 & 59.1 & 31.2 & \first{81.0} & 217.5 \\
Wan2.2~\cite{wan} & 7.7 & 52.0 & 58.4 & 46.2 & 76.5 & 240.8 \\
CogVideoX~\cite{cogvideox} & 1.4 & 68.5 & 67.1 & 22.2 & 75.5 & 234.7 \\
\midrule
\multicolumn{7}{l}{\textit{Camera-controlled models}}\\
Gen3C~\cite{gen3c} & 0.0 & 30.6 & 82.4 & \second{90.6} & 57.6 & 261.2 \\
Hunyuan WorldPlay~\cite{worldplay} & 0.0 & 72.2 & 88.2 & 55.4 & 64.9 & 280.7 \\
Genie 3~\cite{genie3} & 2.9 & 15.2 & 27.3 & 84.7 & \second{78.6} & 208.7 \\
LingBot-World~\cite{lingbotworld} & 3.4 & 40.7 & 76.3 & 35.6 & 67.8 & 223.8 \\
InSpatio-World~\cite{inspatio} & 4.5 & \second{79.6} & \first{99.6} & 0.0 & 69.7 & 253.4 \\
ReMind~\cite{remind} & \second{14.5} & 71.5 & \second{96.4} & 41.6 & 76.9 & \second{300.9} \\
\rowcolor{oursbg}\textbf{\paper (ours)} & \first{17.6} & \first{81.2} & 91.8 & \first{97.6} & 76.5 & \first{364.7} \\
    \bottomrule
  \end{tabular}}
  \vspace{-1em}
\end{table}

\paragraph{STEVO-Bench.}
As shown in \cref{tab:stevo}, \paper achieves the highest Total (364.7, +63.8 over ReMind) and the best State Progress, Physical Plausibility, and Observation Control.
Its Action Control (76.5) is on par with ReMind (76.9), a difference of less than one task.
Coherence trades off against State Progress.
All tasks on which \paper fails Coherence are judged as a ``state jump'' because the subject is in a different state on return, \ie, the state did evolve out of view.
Models whose state barely changes appear more coherent, and as noted in~\cite{remind}, InSpatio-World reaches 99.6 Coherence with 0.0 Observation Control and 4.5 State Progress.
\paper accepts a small Coherence cost in exchange for states that actually evolve and complete out of sight.
Examples are in the supplementary material.

\subsection{Qualitative Results}
\label{sec:exp_qual}

In \cref{fig:qualitative}, InSpatio-World returns to the waiter still holding the tray and cuts the patient off at the frame border, and Gen3C distorts the restaurant while turning away.
Generally, baselines often change the scene while turning away, lose or duplicate the subject, or show it in its initial state on return.
\paper keeps the returned view consistent with the input frame, completes state changes such as sitting down, crouching, or putting an object down, and maintains them on return.
More baselines, examples, and world-state visualizations are in the supplementary material.

\subsection{Ablation Studies}
\label{sec:exp_ablation}

\begin{table}[t]
  \centering
  \caption{\textbf{Training stages of $\E$} on WRBench. All rows use direct projection to isolate $\E$. $\Delta$Avg.: change over the previous row.}
  \vspace{-0.5em}
  \label{tab:stages}
  \setlength{\tabcolsep}{2.4pt}
  \resizebox{\linewidth}{!}{%
  \begin{tabular}{@{}lcccccccc@{}}
    \toprule
    & Cam & & \multicolumn{2}{c}{Visible} & \multicolumn{2}{c}{Re-observed} & & \\
    \cmidrule(lr){4-5}\cmidrule(lr){6-7}
    $\E$ & Align. & Integ. & Spat. & State & Spat. & State & Avg. & $\Delta$Avg. \\
    \midrule
    Zero-shot~\cite{panoworld} & .924 & .645 & .677 & .531 & .687 & .625 & .682 & -- \\
    + Completion & .922 & .661 & .743 & .599 & .740 & .654 & .720 & +.038 \\
    + Evolution & \textbf{.964} & \textbf{.704} & \textbf{.788} & \textbf{.663} & \textbf{.771} & \textbf{.683} & \textbf{.762} & +.042 \\
    \bottomrule
  \end{tabular}}
  \vspace{-0.5em}
\end{table}

\begin{table}[t]
  \centering
  \caption{\textbf{Effect of the \renderer.} Both rows use the same \evolver output, and ``w/o Renderer'' projects the decoded panorama directly. Sharpness: Laplacian variance.}
  \vspace{-0.5em}
  \label{tab:renderer}
  \setlength{\tabcolsep}{4pt}
  \small
  \resizebox{\linewidth}{!}{%
  \begin{tabular}{@{}llccc@{}}
    \toprule
    & Metric & w/o Renderer & \paper & $\Delta$ \\
    \midrule
    \multirow{5}{*}{\rotatebox{90}{WRBench}}
    & Integ. & .704 & .720 & +.016 \\
    & Visible spatial & .788 & .799 & +.011 \\
    & Visible state & .663 & .651 & $-$.012 \\
    & Re-observed state & .683 & .646 & $-$.037 \\
    & Avg. & .762 & .753 & $-$.009 \\
    \midrule
    \multirow{5}{*}{\rotatebox{90}{Quality}}
    & VBench (8-dim mean) & .818 & \textbf{.854} & +.036 \\
    & Imaging quality & .538 & \textbf{.729} & +.191 \\
    & Aesthetic quality & .449 & \textbf{.498} & +.049 \\
    & CLIP-IQA+ & .397 & \textbf{.619} & +.222 \\
    & Sharpness & 35 & \textbf{291} & +256 \\
    \bottomrule
  \end{tabular}}
  \vspace{-1.5em}
\end{table}

\begin{figure}[t]
  \centering
  \includegraphics[width=\linewidth]{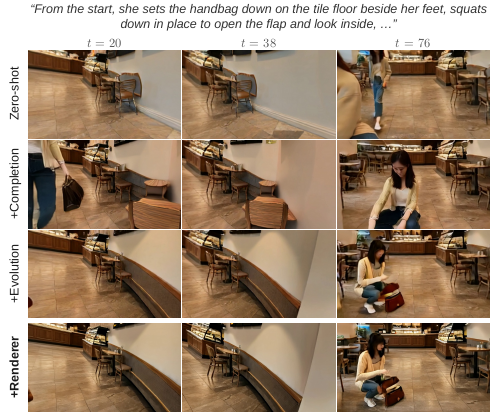}
  \vspace{-1.5em}
  \caption{\textbf{Qualitative ablation.} Each row adds one component (columns: $t{=}20$, 38, 76). Zero-shot, the woman walks past instead of setting her handbag down, the Completion stage has her squat as her bag vanishes, the Evolution stage completes the action out of view, and the Renderer sharpens it without altering the state.}
  \label{fig:qual_ablation}
  \vspace{-1em}
\end{figure}

\paragraph{Training stages of the \evolver.}
Applied zero-shot to a single image, PanoWorld completes the scene but evolves it poorly (.682, \cref{tab:stages}).
The Completion stage improves the official average by $+.038$, mainly through visible spatial and state consistency ($+.066$ and $+.068$), while leaving camera alignment unchanged.
The Evolution stage adds another $+.042$ and improves every metric, including re-observed state ($+.029$) and camera alignment ($+.042$), for an overall gain of $+.080$ over the zero-shot model.
Large-scale completion data teaches the model to continue the scene from a single view, and verified event data teaches the objects in it to evolve, as illustrated in \cref{fig:qual_ablation}.

\paragraph{Effect of the \renderer.}
\Cref{tab:renderer} compares \paper with directly projecting the panorama decoded from the same \evolver output.
Direct projection is blurry (sharpness 35 vs.\ 202 for the input frames), and the \renderer improves every quality measure, raising imaging quality from .538 to .729 and the VBench mean on every scene.
On WRBench, the average changes by only $-.009$.
Integrity and visible spatial consistency improve, while the state metrics decrease slightly, most for re-observed state.
When the rendered videos are downsampled to the native resolution of the panorama, visible and re-observed spatial and state consistency all return to or above the level of direct projection, and re-observed state, which decreased most, even exceeds it (supplementary material).
We therefore attribute the decrease to the sensitivity of the VLM-based judge to appearance rather than to altered states.
The \renderer thus preserves the evolved state while substantially improving visual quality.

\paragraph{Renderer components.}
\Cref{tab:renderer_ablation} removes one reading pathway at a time.
Removing the projected latent maps causes the largest degradation on every metric ($-.024$ Avg.), above all in camera alignment (.950 to .902) and re-observed state ($-.021$).
Without this pixel-aligned estimate, the model follows the camera less reliably, so latent-space projection is the primary reading pathway.
Removing frustum attention lowers the average by .008 and imaging quality by .015.
Every metric is also lower than for the variant without the read-out loss, which isolates the contribution of the attention itself: it retrieves panoramic context that the projection does not provide.
Removing only the read-out loss lowers camera alignment ($-.015$), imaging quality ($-.010$), and every state metric, indicating that the auxiliary supervision helps the attention read the panorama.

\begin{table}[t]
  \centering
  \caption{\textbf{Ablation of the \renderer components.} Each variant removes one component, is trained for 13,500 steps, and is tested on the same $\E$ output. Removing frustum attention also removes the read-out loss. Imaging: VBench imaging quality.}
  \vspace{-0.5em}
  \label{tab:renderer_ablation}
  \setlength{\tabcolsep}{3pt}
  \resizebox{\linewidth}{!}{%
  \begin{tabular}{lcccccc}
    \toprule
    & Cam & Vis. & \multicolumn{2}{c}{Re-observed} & & \\
    \cmidrule(lr){4-5}
    Variant & Align. & State & Spat. & State & Avg. & Imaging \\
    \midrule
    Full & \textbf{.950} & \textbf{.651} & \textbf{.754} & \textbf{.646} & \textbf{.753} & \textbf{.729} \\
    w/o read-out loss & .935 & .647 & .748 & .643 & .751 & .719 \\
    w/o frustum attention & .931 & .645 & .742 & .637 & .745 & .714 \\
    w/o projected latent & .902 & .642 & .732 & .625 & .729 & .703 \\
    \bottomrule
  \end{tabular}}
  \vspace{-0.5em}
\end{table}

\subsection{Efficiency}
\label{sec:exp_cost}

\begin{table}[t]
  \centering
  \caption{\textbf{Inference cost} for one 77-frame view. Parameters and FLOPs are those of the diffusion transformers, excluding encoders and VAEs. FLOPs of $\E$ are amortized over the five camera settings rendered from one panorama in WRBench (unamortized in parentheses). Peak memory is that of diffusion sampling.}
  \vspace{-0.5em}
  \label{tab:cost}
  \setlength{\tabcolsep}{3pt}
  \small
  \resizebox{\linewidth}{!}{%
  \begin{tabular}{lcccc}
    \toprule
    & Params & \makecell{New\\trainable} & TFLOPs & \makecell{Peak\\memory} \\
    \midrule
    \evolver $\E$ & 2.40B & 0.106B & 7,752 (38,759) & 10.5\,GiB \\
    \renderer $\R$ & 1.80B & 0.385B & 27,820 & 8.8\,GiB \\
    \paper & 4.20B & 0.491B & 35,572 (66,579) & 10.5\,GiB \\
    \bottomrule
  \end{tabular}}
  \vspace{-1.5em}
\end{table}

As $\E$ is camera-independent, one panorama serves every trajectory (\cref{tab:cost}).
LoRA adds 1.25\% to the DiT FLOPs of $\E$ and can be merged before inference, and the new modules of $\R$ account for 6.3\% of its FLOPs.
$\E$ and $\R$ run sequentially, so peak memory is that of the larger stage.

\section{Conclusion and Limitations}
\label{sec:conclusion}

We presented \paper, which evolves the world state in the latent space of a panoramic video diffusion model and renders views from it with a perspective video diffusion model.
Trained on event-centric data, the panoramic model evolves object-level states from a single image, and the renderer turns the evolved state into high-quality views.
State-of-the-art results on WRBench and STEVO-Bench suggest that a persistently evolving latent world state is a promising core for video world models, and since the two models share a latent space, either model can be upgraded separately.

\paper handles camera rotation about a fixed center.
Camera translation reveals parallax and newly visible surfaces that a single-center panorama cannot represent, and extending the world state accordingly requires depth, visibility, and multi-view data.
Local physical processes such as melting or liquid-level changes remain rare in real 360$^\circ$ footage and are underrepresented in \data.

{
    \small
    \bibliographystyle{ieeenat_fullname}
    \bibliography{main}
}

\end{document}